\documentclass[lettersize,journal]{IEEEtran}
\usepackage{amsmath,amsfonts}
\usepackage{algorithmic}
\usepackage{array}
\usepackage{textcomp}
\usepackage{stfloats}
\usepackage{url}
\usepackage{verbatim}
\usepackage{graphicx}
\usepackage{cite}
\usepackage{caption}
\usepackage{subcaption}
\usepackage[ruled]{algorithm2e}
\usepackage{cleveref}
\usepackage{mwe}
\begin{document}

\title{Learning-Based Biharmonic Augmentation for Point Cloud Classification}

\author{Jiacheng Wei, Guosheng Lin, Henghui Ding, Jie Hu and Kim-Hui Yap
\thanks{Jiacheng Wei and Kim-Hui Yap are with the School of Electrical and Electronic Engineering, Nanyang Technological University, Singapore.
(E-mail: jiacheng002@e.ntu.edu.sg, ekhyap@ntu.edu.sg) }
\thanks{Guosheng Lin and Henghui Ding are with the School of Computer Science and Engineering, Nanyang Technological University, Singapore. (e-mail:
gslin@ntu.edu.sg).}
\thanks{Jie Hu is with State Key Lab of Computer Science, ISCAS \& University of Chinese Academy of Sciences.}
\thanks{Manuscript received October 23, 2023.}}


\maketitle

\begin{abstract}
Point cloud datasets often suffer from inadequate sample sizes in comparison to image datasets, making data augmentation challenging. While traditional methods, like rigid transformations and scaling, have limited potential in increasing dataset diversity due to their constraints on altering individual sample shapes, we introduce the Biharmonic Augmentation (BA) method. BA is a novel and efficient data augmentation technique that diversifies point cloud data by imposing smooth non-rigid deformations on existing 3D structures. This approach calculates biharmonic coordinates for the deformation function and learns diverse deformation prototypes. Utilizing a CoefNet, our method predicts coefficients to amalgamate these prototypes, ensuring comprehensive deformation. Moreover, we present AdvTune, an advanced online augmentation system that integrates adversarial training. This system synergistically refines the CoefNet and the classification network, facilitating the automated creation of adaptive shape deformations contingent on the learner status. Comprehensive experimental analysis validates the superiority of Biharmonic Augmentation, showcasing notable performance improvements over prevailing point cloud augmentation techniques across varied network designs.
\end{abstract}


\begin{figure}[t]
\centering
\subfloat[Original point cloud sample.]{
  \includegraphics[width=0.99\columnwidth]{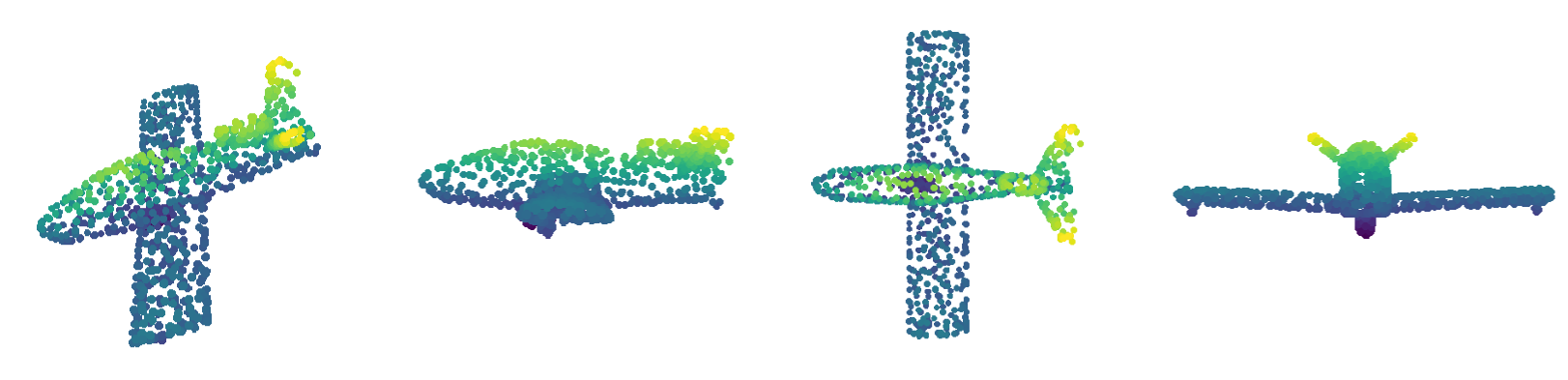}
}

\subfloat[Learned deformation prototypes.]{
  \includegraphics[width=0.99\columnwidth]{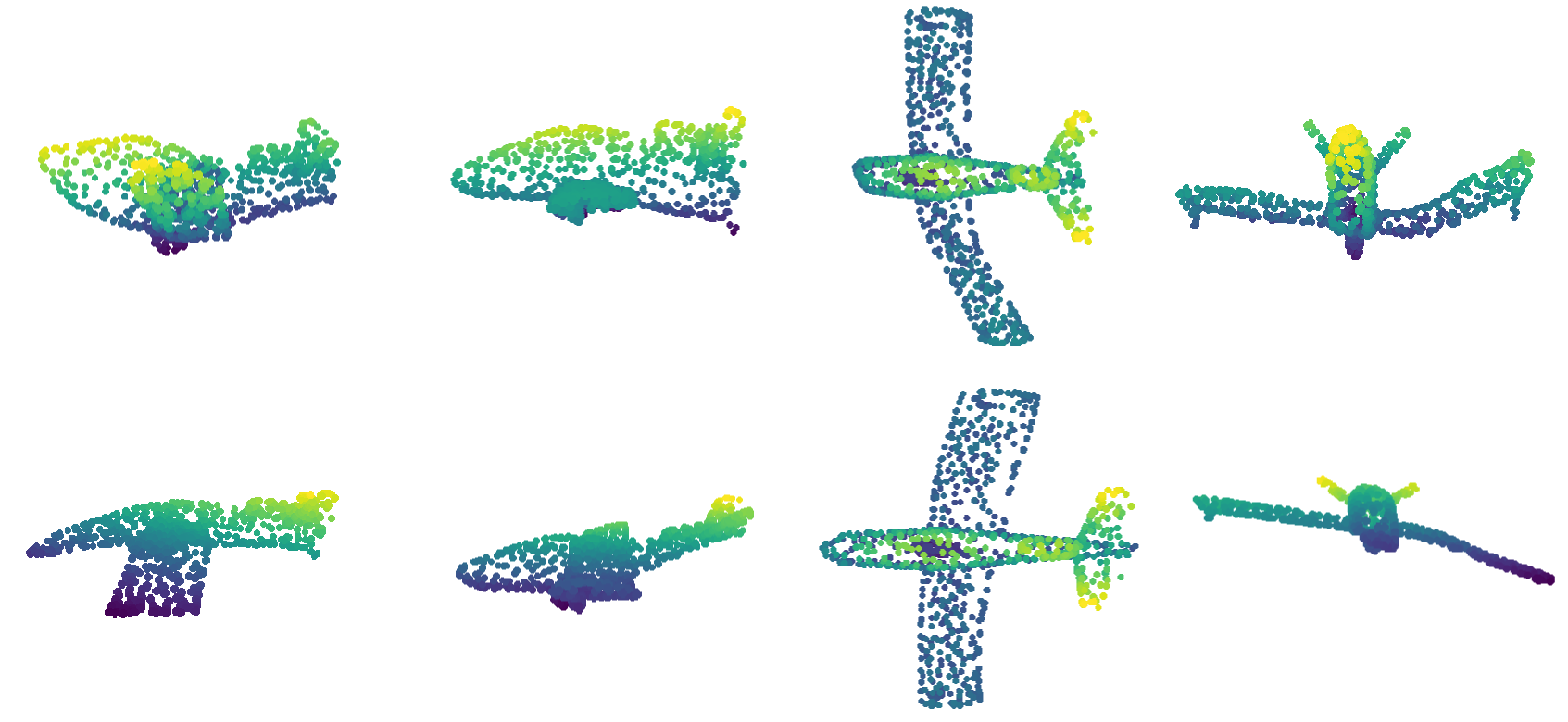}
  
}\label{fig1:metahandle}

\subfloat[Augmented samples.]{
  \includegraphics[width=0.99\columnwidth]{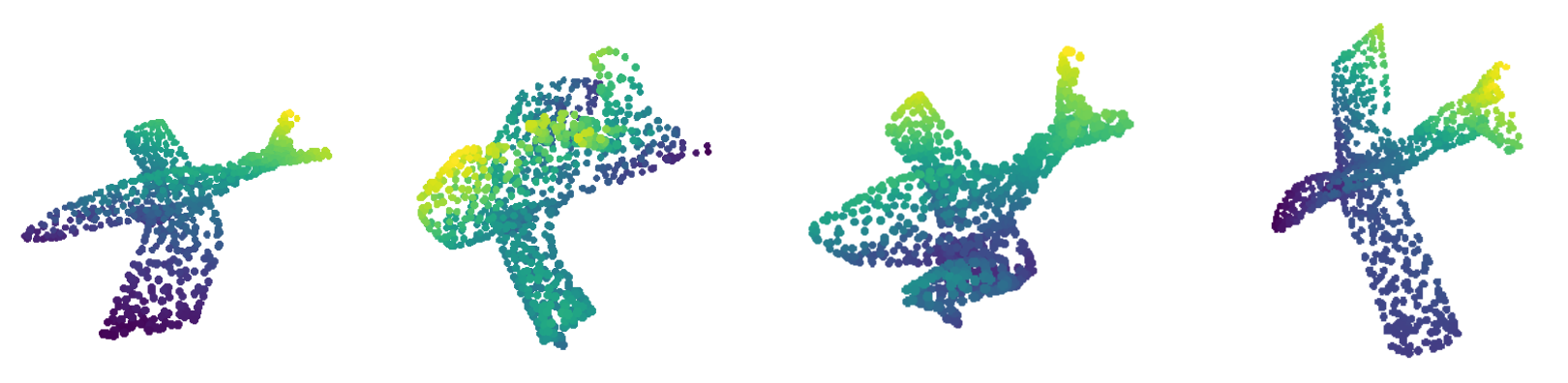}
}

\caption{(a) Original airplane point cloud depicted from three distinct perspectives. (b) Set of deformation prototypes derived from the initial shape, each representing a feasible deformation basis. (c) Augmented samples formed by blending the deformation prototypes. While these samples exhibit notable shape variations compared to the original, their recognition as airplanes remains straightforward.}
\label{fig1}

\vspace{-15pt}
\end{figure}

\section{Introduction}

\IEEEPARstart{O}{v}er recent years, the potential for 3D point cloud recognition has expanded considerably, driven by enhancements in 3D sensor technology. Deep learning methodologies, as evidenced by works such as~\cite{thomas2019kpconv, choy20194d}, have furthered advancements in 3D computer vision. Nonetheless, 3D vision tasks are particularly vulnerable to data limitations compared to their 2D counterparts. Notably, while 2D datasets like ImageNet~\cite{krizhevsky2012imagenet} and COCO~\cite{lin2014microsoft} boast extensive size and diversity, 3D point cloud datasets such as ModelNet40~\cite{wu20153d} are comparatively limited in both aspects. Furthermore, given the rising demand, there exists a noticeable research gap in the development of augmentation techniques for 3D point cloud data.

Traditional augmentation approaches for point cloud data, termed Conventional Data Augmentation (CDA)\cite{qi2017pointnet, qi2017pointnet++}, chiefly encompass translation, scaling, rotation, and point-wise noise addition. PointAugment\cite{li2020pointaugment} brings forth an auto-augmentation framework~\cite{cubuk2019autoaugment}, employing an auxiliary network to produce a global transformation matrix paired with point-wise noise. Inspired by techniques such as Mixup~\cite{zhang2017mixup} and ManifoldMixup~\cite{verma2019manifold}, PointMixup~\cite{chen2020pointmixup} formulates virtual samples by interpolating two samples from discrete classes, both spatially and feature-wise. RSMix~\cite{lee2021regularization}, adopting the CutMix approach~\cite{yun2019cutmix}, constructs virtual samples by substituting portions from alternate samples. Nevertheless, a notable limitation of these strategies is their neglect of modifications to the geometric shape or local structures of individual samples, which might be pivotal for bolstering data diversity.

Two very recent works  explore region-based augmentation for point cloud data. PatchAugment~\cite{sheshappanavar2021patchaugment} employs various Conventional Data Augmentation (CDA) techniques within each ball query~\cite{qi2017pointnet++} localized area. However, this approach can introduce discontinuities, yielding samples that may appear artificial. An inherent limitation is its integration within the ball query, making it unsuitable for networks lacking ball query functionalities.

Conversely, PointWOLF~\cite{kim2021point} applies CDA methodologies on localized areas but incorporates the Nadaraya-Watson kernel regression~\cite{nadaraya1964estimating, watson1964smooth} to refine local transformations. This leads to more authentic-looking augmentations. Yet, the augmentation process entails executing the kernel regression technique $M$ times for each of the $M$ transformed local regions. As these local transformations are generated in isolation, there is a risk of producing incongruent deformations, yielding potentially inauthentic augmentations that might impede effective training. This risk can be mitigated by confining transformations within a meticulously crafted narrow search domain. Hence, the crux of the challenge is the efficient creation of augmentations that embody both plausible and authentic non-rigid deformations.

Thus, the central challenge is to adeptly produce augmentations characterized by both plausible and authentic non-rigid deformations

\begin{figure}
  \centering
  \begin{subfigure}{0.475\linewidth}
    \includegraphics[width=\columnwidth]{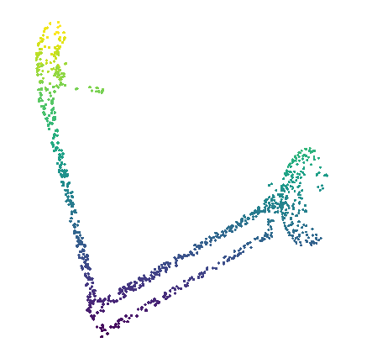}
    \caption{Original.}
    \label{fig:random-a}
  \end{subfigure}
    \hfill
  \begin{subfigure}{0.475\linewidth}
    \includegraphics[width=\textwidth]{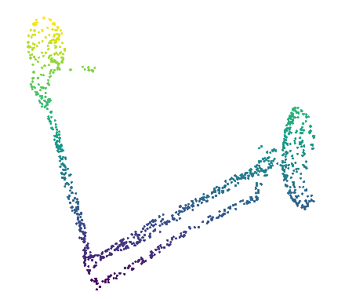}
    \caption{$\sigma=0.01$.}
    \label{fig:random-b}
  \end{subfigure}
  \vskip\baselineskip
  \begin{subfigure}{0.475\linewidth}
    \includegraphics[width=\textwidth]{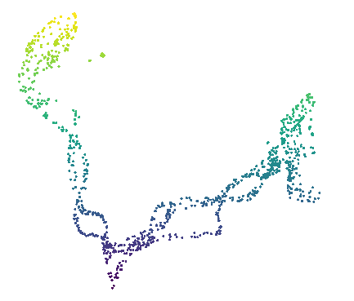}
    \caption{$\sigma=0.05$}
    \label{fig:random-c}
  \end{subfigure}
  \hfill
  \begin{subfigure}{0.475\linewidth}
    \includegraphics[width=\textwidth]{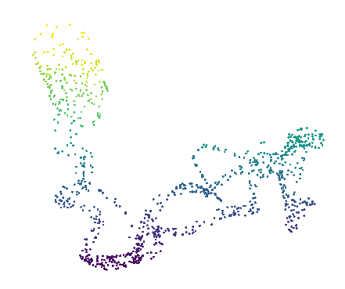}
    \caption{$\sigma=0.1$.}
    \label{fig:random-d}
  \end{subfigure}

  \caption{Deformations of a sample lamp with random offsets $ O \sim \mathcal{N}(0,\,\sigma^{2})$ draw from a normal distribution for each individual control points.}
  \label{fig:random}
  \vspace{-15pt}
\end{figure}

\begin{figure*}[t]
  \centering
    \includegraphics[width=0.99\textwidth]{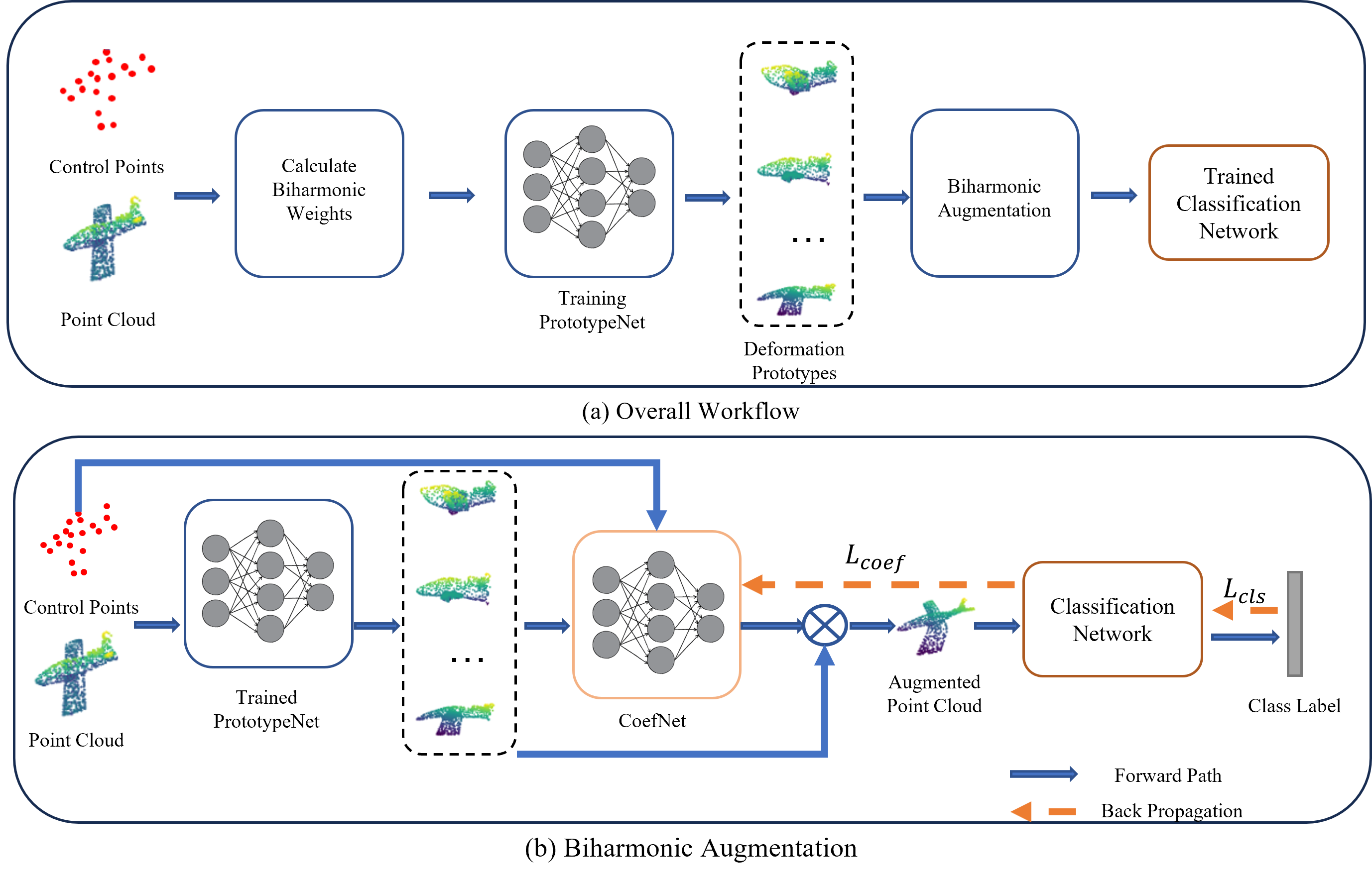}
    \label{fig:ba_framework}
  \caption{The framework of Biharmonic Augmentation. (a) Shows the entire workflow of the proposed method. (b) Shows the details of the Biharmonic Augmentation during the training of the target network. During the adversarial training, the pretrained PrototypeNet is frozen. The PrototypeNet generates deformation prototypes from input point clouds and the sampled control points. The CoefNet takes the deformation prototypes and the control points to generate the combination weights of the prototypes. Then, the augmented point cloud is derived from the linear combination of the prototypes. The CoefNet and the classification network are jointly updated through adversarial training.}
 \vspace{-10pt}
\end{figure*}

In this work, we introduce Biharmonic Augmentation (BA), an innovative point cloud data augmentation technique designed to surmount prevailing challenges and amplify dataset diversity. As depicted in \Cref{fig:ba_framework}, our approach refrains from directly crafting deformation offsets for each data point, either on a local or global scale. Instead, we leverage classical \textit{deformation handles} from the realm of computational geometry to parameterize smooth deformations. These handles consist of a concise set of \textit{control points} extracted from the dataset. We adopt \textit{biharmonic coordinates}~\cite{wang2015linear}, anchored on these control points, as our deformation function. This facilitates the sculpting of the shape by adjusting the coordinates of control points. The inherent advantage of biharmonic coordinates is their capability to yield smooth deformations through straightforward matrix multiplication. Each control point inherently affects the deformation in a localized subspace. Nonetheless, individual modifications to these control points can lead to unnatural deformations. For instance, \Cref{fig:random} presents smooth but incongruous deformations created by superimposing random coordinate offsets on control points of a sample lamp, making its identification as a lamp challenging for human viewers. To counteract this, we derive inspiration from \cite{liu2021deepmetahandles} to cultivate a suite of deformation prototypes. As illustrated in \Cref{fig1:metahandle}, each prototype encompasses a set of coordinate offsets for every control point, epitomizing a viable deformation foundation. By amalgamating all prototypes with diverse coefficients, we can craft a myriad of credible deformations. As a result, our deformation methodology ensures a harmonious balance between localized smoothness and overarching authenticity in the enhanced point clouds.

For optimal augmentations tailored to classification tasks, our goal is to pinpoint target shapes conducive to classification training. This involves determining the most apt coefficients for the deformation prototypes. To achieve this, we introduce CoefNet, designed to predict these coefficients for automated augmentation creation. Instead of directly generating the coefficients, CoefNet predicts offsets for coefficients chosen at random, rectifying any unfavorable deformations. This strategy ensures that CoefNet does not gravitate towards particular deformations, thereby preventing both networks from becoming ensnared in local minima.

Drawing inspiration from notable works~\cite{tang2020onlineaugment, zhang2019adversarial, xie2020adversarial, Goodfellow2015ExplainingAH, Kurakin2017AdversarialML}, we introduce AdvTune, an online augmentation framework. AdvTune seamlessly integrates adversarial learning methodologies, enabling concurrent training of both the CoefNet and classification networks. Consequently, our Biharmonic Augmentation is adept at autonomously producing augmentations that are both smooth and credible, dynamically adapting to the state of the learner network

In summary, our contributions can be summarized as:
\begin{itemize}

\item We present Biharmonic Augmentation, a pioneering data augmentation technique adept at enhancing data diversity by proficiently crafting smooth and authentic non-rigid deformations within point cloud data.
\item We architect AdvTune, an online adversarial auto-augmentation framework, adeptly producing augmentations that are attuned to the current state of the learner network.
\item Through empirical studies, we validate that our Biharmonic Augmentation yields substantial advancements in point cloud shape classification efficacy, surpassing prevailing augmentation approaches across diverse network backbones.

\end{itemize}

\section{Related Works}\label{sec:related}
\textbf{Data Augmentation for 2D Images:} In 2D vision tasks, common random data augmentations~\cite{simard2003best} include flipping, random cropping, and rotation applied to image datasets. Subsequent methods, such as Gaussian noise~\cite{Lopes2019ImprovingRW} and regional dropout~\cite{Devries2017ImprovedRO}, introduce noise or eliminate certain pixels within images. Techniques like Mixup~\cite{zhang2017mixup} employ linear interpolation between two images to craft virtual samples. Advanced approaches, such as ManifoldMixup~\cite{verma2019manifold}, blend two images in the feature domain, while CutMix~\cite{yun2019cutmix} merges images by substituting certain regions from one image into another. AutoAugment~\cite{cubuk2019autoaugment} and its derivatives~\cite{Ho2019PopulationBA, Lee2020LearningAN} integrate AutoML strategies~\cite{Baker2017DesigningNN} to autonomously discern optimal augmentation policies. It is worth noting that these aforementioned strategies are typically classified as \textit{offline} data augmentation techniques, given that the augmentation process remains distinct from the primary model training phase.

\textbf{Adversarial Training for Image Data Augmentation: } DAGAN~\cite{Antoniou2017DataAG} employs the Generative Adversarial Network (GAN)\cite{Goodfellow2014GenerativeAN} to produce augmentations. Works such as\cite{zhang2019adversarial, Goodfellow2015ExplainingAH, Kurakin2017AdversarialML} harness adversarial training as a data augmentation tactic, where the augmented data is infused with adversarial perturbations. While these methods can bolster model robustness, they occasionally compromise generalizability. Efforts like Online Augment~\cite{tang2020onlineaugment} and AdvProp~\cite{xie2020adversarial} seek to enhance generalizability during adversarial training. These techniques are typically categorized as \textit{online} data augmentation methods, given their ability to dynamically generate augmented data in accordance with the evolving state of the training model.

\textbf{Deep Learning on 3D Point Clouds:} While 2D image data is structured with grid pixels, 3D point cloud data consists of unordered sets. This distinction complicates the direct application of established convolutional neural network architectures to point cloud datasets. PointNet~\cite{qi2017pointnet} offers a solution by leveraging shared multi-layer perceptions for individual points and subsequently deriving a global feature through a max-pooling operation. Its enhanced successor, PointNet++\cite{qi2017pointnet++}, utilizes a hierarchical framework, integrating multiple PointNet-inspired architectures to capture and group local data attributes. Several studies~\cite{wang2019dynamic, thomas2019kpconv} have applied convolution operations to point clouds, focusing on nearest neighbors. On the other hand, voxel-based techniques~\cite{choy20194d} segment point cloud data into discrete voxels, conducting sparse convolutions on these segmented units. In the realm of neural networks and learning systems, deep learning methods on point clouds are drawing more attentions, \cite{9750402, 10054469} focus on the point cloud classification task using attention modules.  \cite{9733414, 9653732} proposed graph neural network based methods for point cloud segmentation tasks. \cite{9950286} also utilizes point cloud deformation for reconstruction tasks.

\textbf{Data Augmentation for 3D point clouds:} Conventional data augmentation (CDA)~\cite{qi2017pointnet, qi2017pointnet} applies translation, scaling, rotation and point-wise noise to point cloud data. PointAugment~\cite{li2020pointaugment} introduces an auto-augmentation~\cite{cubuk2019autoaugment} framework to learn to generate a global transformation matrix and point-wise noise with another network. Mixup based methods~\cite{chen2020pointmixup, lee2021regularization} create virtual samples by mixing two point cloud samples. PatchAugment~\cite{sheshappanavar2021patchaugment} and PointWOLF~\cite{kim2021point} performs different conventional data augmentation techniques to different local regions in the point cloud samples. 

\section{Method}
\subsection{Biharmonic Deformation}

The primary objective of our method is to introduce smooth and plausible deformations of existing point cloud data, thereby enriching data diversity. Traditional shape deformation tasks typically involve constraints, such as target shapes, which may limit flexibility. To overcome this limitation, we adopt handle-based deformation, a well-established technique in computational geometry. Unlike directly manipulating individual point positions, handle-based deformation offers a lower degree of freedom, making it more manageable and allowing for smoother and more realistic augmentations. The deformation functions are derived by solving biharmonic equations, leveraging selected handles and their corresponding boundary conditions, as demonstrated in prior studies by Jacobson et al.~\cite{Jacobson2014BoundedBW, Jacobson2012SmoothSF}. This approach empowers our method to create augmented point cloud data with enhanced realism and versatility.

We adopt the efficient solution proposed by Wang et al.~\cite{Wang2015LinearSD} to compute the matrix $\mathbf{W}$, which involves solving biharmonic functions, leading to the derivation of \textit{biharmonic coordinates}. These coordinates enable us to represent the deformation function as $f(\mathbf{C}) = \mathbf{W}\mathbf{C}$, where $\mathbf{C}$ corresponds to the coordinates of the control points. Consequently, with the coordinates of the control points $f(\mathbf{C})$, we can effortlessly reconstruct the original point cloud by employing simple matrix multiplication. This methodology allows us to efficiently and accurately generate smooth and plausible deformations, enhancing the data diversity and realism of the augmented point cloud samples.

In this context, each control point corresponds to the shape of a local region. Deformation can be achieved by adding position offsets $\mathbf{O}\in\mathbb{R}^{c \times 3}$ to the control points, indicating the displacements of the control points. Then, the new control point coordinates will be $\mathbf{C}=\mathbf{C_0}+\mathbf{O}$, where $\mathbf{C_0} \in\mathbb{R}^{c \times 3}$ represents the original rest position of the control points, and the deformation function can be rewritten as
\begin{equation}
f(\mathbf{O}) = \mathbf{W}(\mathbf{C_0}+\mathbf{O}),
\label{eq:f}
\end{equation}
The deformation function has $3c$ degrees of freedom. As demonstrated in \Cref{fig:random}, we can generate smooth deformations by moving the control points individually, but this may result in implausible deformations that contradict the inherent characteristics of the original shapes.

\begin{figure*}[t]
  \centering
    \includegraphics[width=0.95\textwidth]{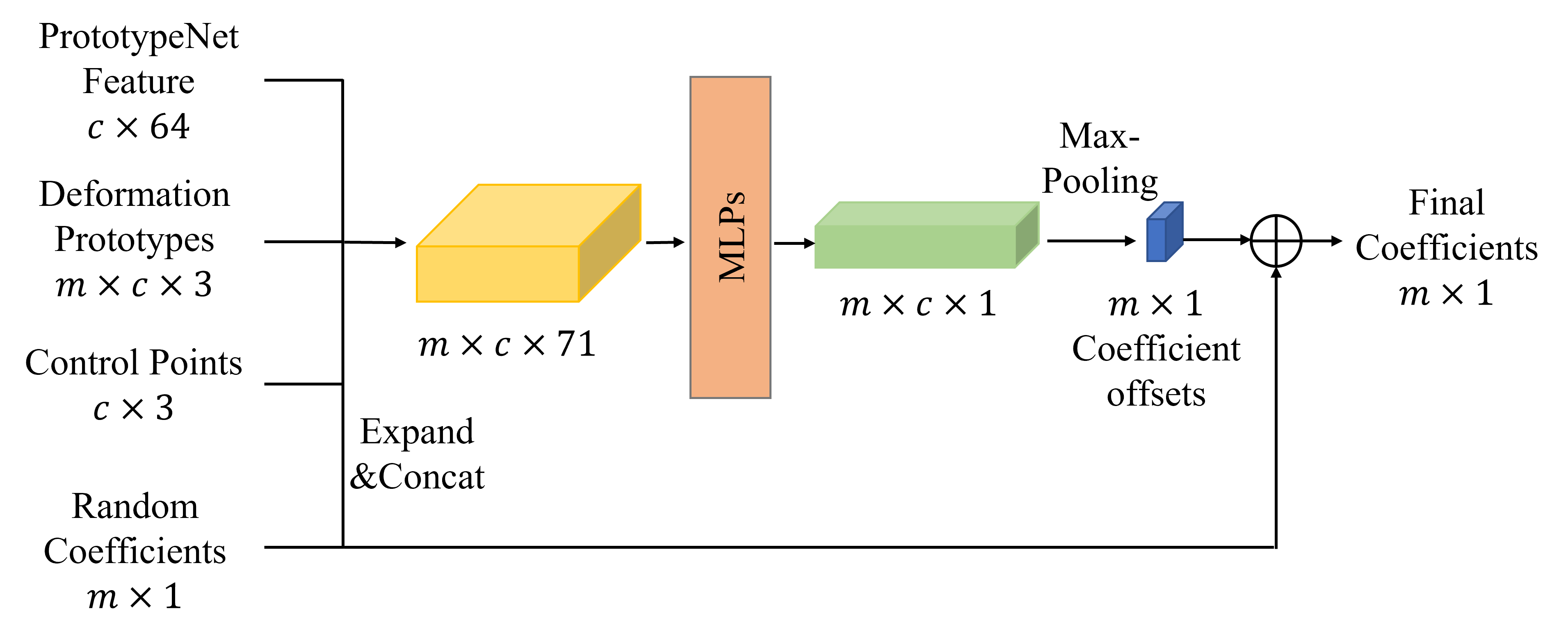}

    \label{fig:coef}
  \caption{Our CoefNet takes the PrototypeNet features, the learned deformation prototypes, the control point rest positions and a set of random coefficient sampled from a normal distribution and generates a coefficient offsets to correct the random coefficients.}
 \vspace{-20pt}
\end{figure*}

\subsection{Deformation Prototypes}

We adopt the approach presented in \cite{liu2021deepmetahandles} to learn a set of deformation prototypes that impose constraints and correlations on the control points in a lower-dimensional subspace. For each point cloud, we generate $m$ deformation prototypes $\mathbf{M}_i \in \mathbb{R}^{c \times 3}$. Each deformation prototype contains a set of offsets for all the control points:

\begin{equation}
   \mathbf{M}_i = [\vec{o}_{i1}, \cdots, \vec{o}_{ic}]^T,
\end{equation}

where $\vec{o}_{ij} \in \mathbb{R}^3$ is the offset of the $j$-th control point in deformation prototypes $\mathbf{M}_i$. Each deformation prototype indicates a plausible global deformation direction for the point cloud sample. By introducing a set of combination coefficients $\mathbf{a} = [a_1, \cdots, a_m]$, the overall control point offsets can be derived by a linear combination of the deformation prototypes: $\mathbf{O} =\sum_{i=1}^{m}a_i\mathbf{M}_i$. Then, the new deformation function is defined as:

\begin{equation}
  g \left( \mathbf{a}; \{ \mathbf{M}_i \}_{i=1}^m \right) = \mathbf{W} (\mathbf{C_0} + \sum_{i=1}^{m}a_i\mathbf{M}_i),
  \label{eq:g}
\end{equation}

The new deformation function $g: \mathbb{R}^{m} \rightarrow \mathbb{R}^{n \times 3}$ thus has only $m$ degrees of freedom. However, with different combination coefficients, we can generate arbitrary many plausible deformations. 

We generate the deformation prototypes by adopting Meta-HandleNet proposed by~\cite{liu2021deepmetahandles} and denote it PrototypeNet. The PrototypeNet is trained by deforming a source point cloud to a target point cloud sample in the same category. We adopt most of the training strategy from ~\cite{liu2021deepmetahandles} with some modifications to adapt to the shape classification task. Instead of training separate models for each category, we train a single model for all the classes for efficiency. Due to the limited space, we will detail the training process of PrototypeNet in supplementary materials. Since the target samples are all other samples from the same category, the deformation prototypes are learned to factorize the subspace that covers all the plausible deformations. To improve the efficiency in the later data augmentation process, we save the generated deformation prototypes and an intermediate feature for each point cloud training sample for later usage. 

We use the biharmonic deformation as a tool to generate augmentations. However, shape deformation methods require a target shape as input. Therefore, the problem lies in finding the optimal target shapes, which are the augmented shapes that effectively help the classification training.

\subsection{CoefNet}\label{sec:coefnet}

The naive way to perform the data augmentation is to draw a set of random coefficients from a normal distribution. However, as discussed in \Cref{sec:related}, many previous works show that random augmentation is less efficient and effective than searching or learning-based augmentation methods. Hence, we design a CoefNet to automatically generate the combination coefficients $\mathbf{a}$. Meanwhile, to prevent the CoefNet from converging to some local minima and producing constant deformations. It is still important to bring randomness into the network. A conventional way is to feed random noise along with the inputs to the network. However, the generated coefficients may not provide enough diversity in the augmentation. We combine the above two methods and design our CoefNet to predict offsets to the randomly sampled coefficients. The randomly sampled coefficients provide sufficient diversity to the augmentation while the predicted offsets can correct unwanted deformation and guide the augmentation to adapt to the learner state.

Specifically, as shown in \Cref{fig:coef}, given a random coefficient $\mathbf{a^{rand}} = [a_1^{rand}, \cdots, a_m^{rand}] \sim \mathcal{N}(\mu_a,\,\sigma_a^{2})$, we expand it by $c$ times, where $c$ is the number of control points. Similarly, we expand the control points coordinates and the stored PrototypeNet features $\mathbf{F_{mh}}$ by $m$ times. Then, we concatenate these inputs together with the generated deformation prototypes and form an overall feature size $m \times c \times 71$. Then, we feed this feature to three MLP layers and generate a $m \times c \times 1$ tensor. We apply max-pooling on the tensor along the $c$ dimension to get the coefficient offsets $\mathbf{a^{off}}$. The overall coefficients is derived by $\mathbf{a} = \mathbf{a^{rand}} + \beta\mathbf{a^{off}}$.

\Cref{algorithm1} summarizes the whole process to generate deformed augmentation samples.

\begin{figure*}[t]
  \centering
    \includegraphics[width=0.99\linewidth]{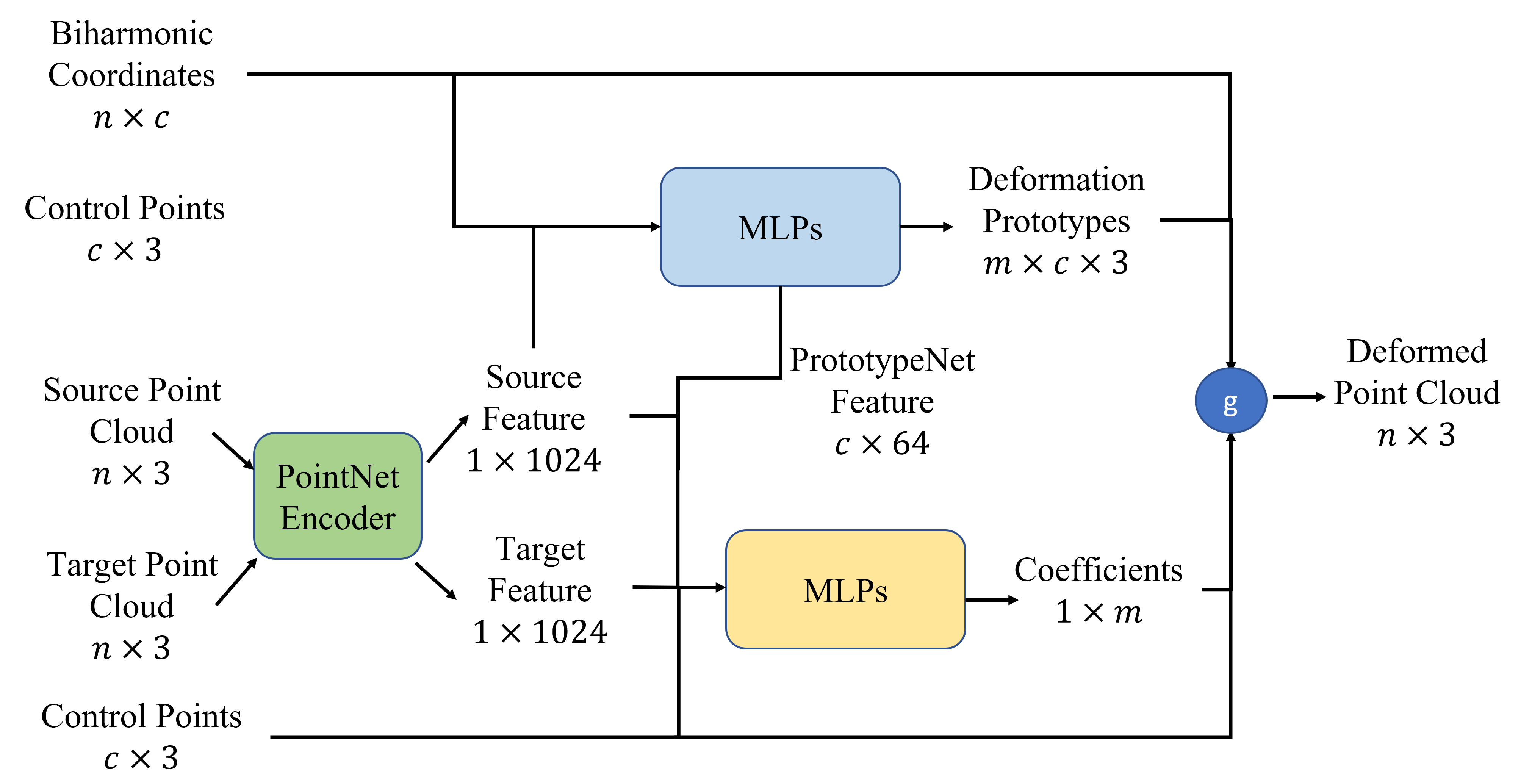}

  \caption{The network architecture of PrototypeNet during training, and we only save the generated deformation prototypes and the intermediate feature for our further process. Here $g$ is the deformation function in \Cref{eq:g}.}
  \label{fig:metatrain}
 \vspace{-15pt}
\end{figure*}

\subsection{AdvTune}
In the following section, we present AdvTune, an online data augmentation framework with adversarial training, designed to jointly optimize the classification networks and CoefNet. This process guides the CoefNet to generate deformations that best fit the classification network.

The original purpose of adversarial training~\cite{Madry2018TowardsDL} is to enhance model robustness against adversarial attacks. However, such trained models exhibit poor generality on clean or original samples~\cite{Xie2019FeatureDF, Madry2018TowardsDL}. As discussed by \cite{xie2020adversarial}, the problem is caused by a \textit{distribution mismatch}, where the original samples and adversarial samples lie on two different distributions. Adversarial training has been proven useful\cite{xie2020adversarial, tang2020onlineaugment} in assisting network training to achieve both robustness and generality by utilizing its regularization ability, provided that the \textit{distribution mismatch} is well managed.

We propose two strategies to overcome the \textit{distribution mismatch}. The first strategy involves applying regularizations on the augmentor CoefNet during training, preventing excessive distribution displacement from the original data. The second strategy is an iterative training scheme.

\begin{algorithm}[t] 
\DontPrintSemicolon
\KwData{control points $\mathbf{C_0}\in\mathbb{R}^{c \times 3}$,\\
 biharmonic coordinates $\mathbf{W}\in\mathbb{R}^{n \times c}$,\\
 deformation prototypes $\mathbf{M}_{i=0}^m \in \mathbb{R}^{m \times c \times 3}$,\\
 CoefNet $h(.;\mathbf{F_{mh}}, \mathbf{C}, \mathbf{W})$\\}
\KwResult{ augmented point cloud $\mathbf{P^d} \in \mathbb{R}^{n \times 3}$\\}

Sample random coefficients $\mathbf{a^{rand}} = [a_1^{rand}, \cdots, a_m^{rand}] \sim \mathcal{N}(\mu_a,\,\sigma_a^{2})$\\
Update guided coefficients $\mathbf{a} \leftarrow h(\mathbf{a^{rand}};\mathbf{F_{mh}}, \mathbf{C}, \mathbf{W})$ \\
Generate augmentation  $\mathbf{P^d} = \mathbf{W} (\mathbf{C_0} + \sum_{i=1}^{m}a_i\mathbf{M}_i)$.\\
\KwRet{$\mathbf{P^d}$}
\caption{ \textbf{Biharmonic Augmentation}}
\label{algorithm1}
\end{algorithm}

We denote the parameters in the classification network and CoefNet as $\theta$ and $\phi$. Given a point cloud dataset $\{\mathbf{P}_k\}_{k=1}^K$ with $K$ samples, for each sample $\mathbf{P}_k$, a biharmonic coordinates $\mathbf{W}_k$, a set of control points $\mathbf{C_0}_k$ and deformation prototypes $\mathbf{M}_k$. A recovered sample $\mathbf{P^r}_k$ can be derive as $\mathbf{P^r}_k = \mathbf{W}_k\mathbf{C_0}_k$ where $\mathbf{P^r}_k$ is very similar to $\mathbf{P}_k$. Then, we generate a set of coefficients $\mathbf{a}_k$ using CoefNet and produce an augmented sample $\mathbf{P^d}_k = \mathbf{W}_k (\mathbf{C_0}_k + \sum_{i=1}^{m}a_{ki}\mathbf{M}_{ki})$.  

 ~\cite{Goodfellow2015ExplainingAH,Kurakin2017AdversarialML} suggest training the classification network with a mixture of the original and augmented data. We use stochastic gradient descent for illustration. For each mini-batch, $\theta$ is updated by gradients:
 \begin{equation}\label{eq:mixture}
    \nabla_{\theta}\mathcal{L}(\mathbf{P}, y;\theta) + \nabla_{\theta}\mathcal{L}(\mathbf{P^r}, y;\theta)
    + \nabla_{\theta}\mathcal{L}(\mathbf{P^d}, y;\theta, \phi),
\end{equation}
where $\mathcal{L}$ is the loss function, in our case, a cross-entropy loss and $y$ is the ground truth label.

We apply two regularization terms in optimizing the $\phi$ in CoefNet to prevent overwhelming distribution displacement. The first term is a Chamfer Distance~\cite{Fan2017APS} between the recovered and augmented samples.
\begin{align}
\mathcal{R}_{\texttt{CD}}(\mathbf{P^d}, \mathbf{P^r}) = & \sum_{p_1\in \mathbf{P^d}} \min_{p_2 \in \mathbf{P^r}} \|p_1 - p_2\|_2^2 \nonumber \\ 
+ & \sum_{p_2 \in \mathbf{P^r}} \min_{p_1 \in \mathbf{P^d}} \| p_1 - p_2 \|_2^2,
\end{align}
The second term is a symmetry loss~\cite{Wang20193DN3D} of the augmented sample since most of the objects are symmetric in certain directions. Since the shape classification datasets are usually unaligned, we developed an orientation invariant symmetry loss by taking the minimum value of $v$ symmetry loss calculated along $v$ directions separated by $\pi/v$:  
\begin{equation}
    \mathcal{R}_{\texttt{sym}}(\mathbf{P^d}) = \min_{\delta \in \mathbf{\Delta}} \mathcal{R}_{CD}(Mir(Rot(\mathbf{P^d}, \delta)), Rot(\mathbf{P^d}, \delta)),
\end{equation}
where $Mir$ takes the mirror of the point cloud along x-axis, $Rot$ is a rotation function along the up-axis, $\Delta$ is the set of $v$ rotation angles. With these two regularization terms, $\phi$ is updated by:
\begin{equation}\label{eq:aug_update}
\begin{split}
        \lambda_{cd}\nabla_{\phi}\mathcal{R}_{cd}(\mathbf{P^r}, &\mathbf{P^d}; \phi)\\ + \lambda_{sym}\nabla_{\phi}\mathcal{R}_{sym}(\mathbf{P^d};\phi)
    - &\lambda_{adv}\nabla_{\phi}\mathcal{L}(\mathbf{P^d}, y;\theta, \phi),
\end{split}
\end{equation}
where $-\mathcal{L}(\mathbf{P^d}, y;\theta, \phi)$ is the adversarial loss from the classification network and $\lambda$s are the balancing weights.

Fine-tuning is proved to be another solution to relief \textit{distribution mismatch} by ~\cite{xie2020adversarial}. However, a final fine-tuning following the whole training process may not be strong enough to recover all the performance degradation caused by \textit{distribution mismatch}, especially for small-scale networks. We enhance the ability to finetune to neutralize \textit{distribution mismatch} with a simple yet effective iterative training scheme.

\Cref{algorithm2} summarizes the procedure of AdvTune. We separate the classification network training by training with the mixture of $\mathbf{P^r}$ and $\mathbf{P^d}$ as the adversarial learning step and training with $\mathbf{P}$ as the fine-tuning step. Then, \Cref{eq:mixture} is divided into two parts:
\begin{equation}\label{eq:mixture1}
    \nabla_{\theta}\mathcal{L}(\mathbf{P^r}, y;\theta)
    + \nabla_{\theta}\mathcal{L}(\mathbf{P^d}, y;\theta, \phi),
\end{equation}
\begin{equation}\label{eq:mixture2}
    \nabla_{\theta}\mathcal{L}(\mathbf{P}, y;\theta) .
\end{equation}

We establish an inner loop consisting of $E_i$ iterations and an outer loop with $E_o$ iterations. For each iteration in the outer loop, we initially update $\theta$ using \Cref{eq:mixture1} for $E_i$ iterations. Subsequently, $\theta$ is updated with \Cref{eq:mixture2} for an additional $E_i$ iteration to counteract distribution displacement, a process we term distribution tuning. Following this, we update $\phi$ employing \Cref{eq:aug_update}, taking into account the classification network's state with the adversarial loss. Ultimately, after completing $E_o$ iterations of the outer loop, we perform an additional $E_f$ iterations, updating $\theta$ using \Cref{eq:mixture2} as a final fine-tuning step to further mitigate the effects of \textit{distribution mismatch}.

\begin{algorithm}[h!]\small
\caption{\textbf{AdvTune}}
\label{algorithm2}
\KwData{control points $\mathbf{C_0}$, biharmonic coordinates $\mathbf{W}$, deformation prototypes $\mathbf{M}_{i=0}^m$, original point cloud $\mathbf{P}$, Initialized $\theta$ and $\phi$ from classification network and CoefNet}

\KwResult{Updated $\theta$ and $\phi$ \\}
\For{each outer loop}{
\For{each inner loop}{
Generate $\mathbf{P^r} = \mathbf{W}\mathbf{C_0}$ \\
Generate $\mathbf{P^d}$ using \Cref{algorithm1}\\
Update $\theta$ using \Cref{eq:mixture1}
}
\For{each inner loop}{
Update $\theta$ using \Cref{eq:mixture2}
}
\For{each inner loop}{
Generate $\mathbf{P^r} = \mathbf{W}\mathbf{C_0}$ \\
Generate $\mathbf{P^d}$ using \Cref{algorithm1}\\
Update $\phi$ using \Cref{eq:aug_update}
}
}
\For{Final fine-tuning iterations}{
Update $\theta$ using \Cref{eq:mixture2}
}
\KwRet{$\theta$, $\phi$}
\end{algorithm}

\begin{table*}[t]
\centering
\resizebox{0.99\textwidth}{!}{%
\begin{tabular}{c|cccccc|c}
\hline
Model      & CDA~\cite{qi2017pointnet}  & CDA(w/o R) & PointAugment~\cite{li2020pointaugment} & PointMixup~\cite{chen2020pointmixup} & RSMix~\cite{lee2021regularization} & PointWOLF~\cite{kim2021point} & BA(Ours) \\ \hline \hline
PointNet   & 89.2 & 89.7       & 90.9         & 89.9       & 88.7  & 91.1      & \textbf{91.6}   \\
PointNet++ & 90.8 & 92.5       & 92.5         & 91.7       & 91.6  & 93.2      & \textbf{93.3}   \\
DGCNN      & 91.9  & 92.7       & 92.9         & 93.1       & 93.2  & 93.2      & \textbf{93.4}   \\ \hline
\end{tabular}
}
\caption{The overall accuracy on ModelNet40.}
\label{tab:modelnet40}
\end{table*}

\begin{table}[]
\centering
\begin{tabular}{c|cc|c}
\hline
model      & CDA  & PointAugment & BA(Ours) \\ \hline \hline
PointNet   & 91.9 & 94.1         & 94.3            \\
PointNet++ & 93.3 & 94.9         & 95.5            \\
DGCNN      & 94.8 & 95.1         & 95.6            \\ \hline
\end{tabular}%
\caption{The overall accuracy on ModelNet10.}
\label{tab:modelnet10}
\end{table}

\section{Experiments}

\subsection{Datasets} 
We conduct object classification experiments on the widely recognized ModelNet40~\cite{wu20153d} dataset and its subset, ModelNet10. ModelNet40 comprises 9,843 training samples and 2,468 testing samples across 40 classes, generated from CAD models. ModelNet10, a subset of ModelNet40, consists of 10 classes with aligned orientations, containing 3,991 training samples and 908 testing samples. For ModelNet40, we compute the biharmonic coordinates utilizing the raw dataset, but assess our model using the test set from the extensively adopted preprocessed data provided by PointNet~\cite{qi2017pointnet} to ensure a fair comparison. For each sample, we employ 1,024 points as input data, disregarding the surface normal information.

\subsection{Implementation}
\textbf{Backbone Networks.} Our method is model agnostic and we conduct experiments on three popular networks for point clouds, namely PointNet~\cite{qi2017pointnet}, PointNet++~\cite{qi2017pointnet++} and DGCNN~\cite{dgcnn}.

\textbf{Data Preparation}
To compute the biharmonic coordinates on ModelNet40. The process begins by establishing watertight manifolds through the utilization of the method proposed by Huang et al. \cite{huang2018robust} on the raw ModelNet40 data. Following this step, both tetrahedral meshes and surface meshes are generated from the aforementioned manifold, employing the fTetWild algorithm \cite{10.1145/3386569.3392385}. Subsequently, the biharmonic coordinates are computed on the resulting surface mesh using the libigl library \cite{Jacobson2017libiglPG}.

\textbf{Training BA on ModelNet.}  All neural networks and experiments are implemented using PyTorch. We employ the trained PrototypeNet to generate deformation prototypes. For all experiments, the CoefNet is trained with the Adam optimizer, with an initial learning rate of 0.001. The parameter $\beta$ is empirically set at 0.1, and $v$ is selected as 2. The balancing weights $\lambda_{cd}$, $\lambda_{sym}$, and $\lambda_{adv}$ are empirically set at 0.1, 0.1, and 0.01, respectively. The mean $\mu_a$ is set to 0, and the standard deviation $\sigma_a$ is set at 0.5. For PointNet and PointNet++, we use the implementation from ~\cite{Pytorch_Pointnet_Pointnet2}. For DGCNN, we employ its official Pytorch implementation. We adhere to most of the training specifications from the implementations, and we train each model with a batch size of 32. The outer iteration is set to 15, and the inner iteration is set to 10, with each inner iteration corresponding to 1 epoch. The final fine-tuning epoch number is 100.
 
\textbf{Training of PrototypeNet.} Our PrototypeNet is adapted from the MetaHandleNet of \cite{liu2021deepmetahandles} and follows most of the training strategies. \Cref{fig:metatrain} illustrates the architecture of PrototypeNet during training. Unlike \cite{liu2021deepmetahandles}, which trains a model for each class, we train a single network for all classes. Each step takes a source point cloud and a target point cloud from the same category. The two-point clouds are fed into a PointNet\cite{qi2017pointnet} encoder to generate global features. Subsequently, the source feature with control point handles and biharmonic coordinates is concatenated and fed into a group of MLPs to generate PrototypeNet features and predict deformation prototypes with additional MLP layers. Concurrently, the source and target features with control point coordinates and the PrototypeNet features are concatenated and fed into another group of MLPs to generate coefficients that deform the source to the target. Finally, the deformed point cloud is generated using \Cref{eq:g}. We employ the same loss functions $\mathcal{L}_{fit}$, $\mathcal{L}_{symm}$, $\mathcal{L}_{nor}$, $\mathcal{L}_{Lap}$, $\mathcal{L}_{sp}$, $\mathcal{L}_{cov}$, $\mathcal{L}_{ortho}$, and $\mathcal{L}_{SVD}$ to train the network. $\mathcal{L}_{fit}$ and $\mathcal{L}_{symm}$ represent the Chamfer Distance and symmetry loss calculated between the deformed point cloud and the target point cloud. $\mathcal{L}_{nor}$ is the surface normal loss, and $\mathcal{L}_{Lap}$ is the mesh Laplacian loss; these two losses are calculated on the reconstructed surface meshes of the deformed and target point clouds. $\mathcal{L}_{sp}$, $\mathcal{L}_{cov}$, $\mathcal{L}_{ortho}$, and $\mathcal{L}_{SVD}$ are calculated on the deformation prototypes to encourage disentanglement of the deformation prototypes, enabling the deformation prototypes to correspond to different deformation directions. Please refer to \cite{liu2021deepmetahandles} for the details of the loss functions.

The PrototypeNet is designed for shape deformation datasets, which are aligned datasets that all the samples have the same orientation. However, the data are not always aligned in shape classification datasets. Therefore, during the training of PrototypeNet, we rotate the target point cloud along up-axis $v$ times for $2\pi/v$ degrees of each rotation and we calculate the Chamfer Distance between the $v$ rotations of target point cloud with the source point cloud. Then, we take the rotation with minimum Chamfer Distance to the input point cloud as the input target point cloud to eliminate the orientation problem for shape classification datasets. For ModelNet40, we use $v=4$. For each training epoch, we select 20 target point clouds for each source point cloud. Since we train all the classes in a single network, we train the network for 60 epochs instead of the default 30 epochs. We use an Adam optimizer with an initial learning rate of 0.001.

\subsection{Shape Classification}
We compare our proposed BA with several data augmentation methods on point cloud data. CDA denotes conventional data augmentation including global translation, scaling, rotation along up-axis and point-wise jittering. Similar to ~\cite{kim2021point}, we found that rotation along up-axis could be harmful to the network training. Therefore, we also provide the result of CDA without rotation. For fair comparison, we reproduce the results of other methods using their official implementations with the same backbone implementation. PointAugment~\cite{li2020pointaugment} and PointMix~\cite{chen2020pointmixup} are two poineer works in the area. RSMix~\cite{lee2021regularization} and PointWolF~\cite{kim2021point} are two recent works with good performance. 

Our method produces the largest performance gain on all three backbone networks.

\begin{table}[b]

\centering
\begin{tabular}{c|cccccc}
\hline
$\sigma$        & 0    & 0.01 & 0.02 & 0.05 & 0.1  & 0.2  \\\hline
accuracy & 88.0 & 86.9 & 86.9 & 86.8 & 87.2 & 86.9 \\ \hline
\end{tabular}%
\caption{Moving individual control points with random offsets draw from a normal distribution with different standard deviation ($\mu = 0$), no convectional data augmentation during training.}
\label{tab:rand_c}
\end{table}

\subsection{Experiments on Deformation Prototypes}\label{sec:mhexp}

In \Cref{tab:rand_c}, several PointNet models are trained utilizing augmentations generated by adding random offsets to individual control points, while excluding conventional data augmentation techniques. The offsets are derived from normal distributions with $\sigma = (0.01, 0.02, 0.05, 0.1, 0.2)$. When compared to the baseline result of 88.0, all models with random offsets exhibit a decline in classification performance. Additionally, experiments were conducted to train PointNet with randomly generated combination coefficients for deformation prototypes, as shown in \Cref{tab:rand_a}. This can be perceived as a comparison with \cite{liu2021deepmetahandles} when random target shapes are employed. Evidently, deformations generated by random coefficients can enhance network performance in classification, provided they are sampled within a reasonable range ($\sigma_a \le 0.5$). Moreover, as depicted in \Cref{fig:random}, implausible and unrealistic deformation is introduced by individual offsets applied to control points. Consequently, it can be concluded that unrealistic augmentation negatively impacts network training, while deformation prototypes address this issue as they are learned to represent plausible deformation directions.

\begin{table}[]
\centering
\begin{tabular}{c|cccccc}
\hline
$\sigma_a$                              & 0.1                      & 0.2                      & 0.3  & 0.5                      & 0.7  & 1                        \\ \hline
random coefficients            & 89.3                     & 89.6                     & 89.5    & 90.1                     & 87.8    & 78.4                    \\
\multicolumn{1}{l|}{BA} & \multicolumn{1}{l}{90.3} & \multicolumn{1}{l}{90.9} & 91.1 & \multicolumn{1}{l}{91.6} & 91.1 & \multicolumn{1}{l}{90.8} \\ \hline
\end{tabular}%
\caption{BA comparing to randomly generated combination coefficients.}
\label{tab:rand_a}
\end{table}

\begin{figure*}[t]
  \centering
    
    \includegraphics[width=0.9\textwidth]{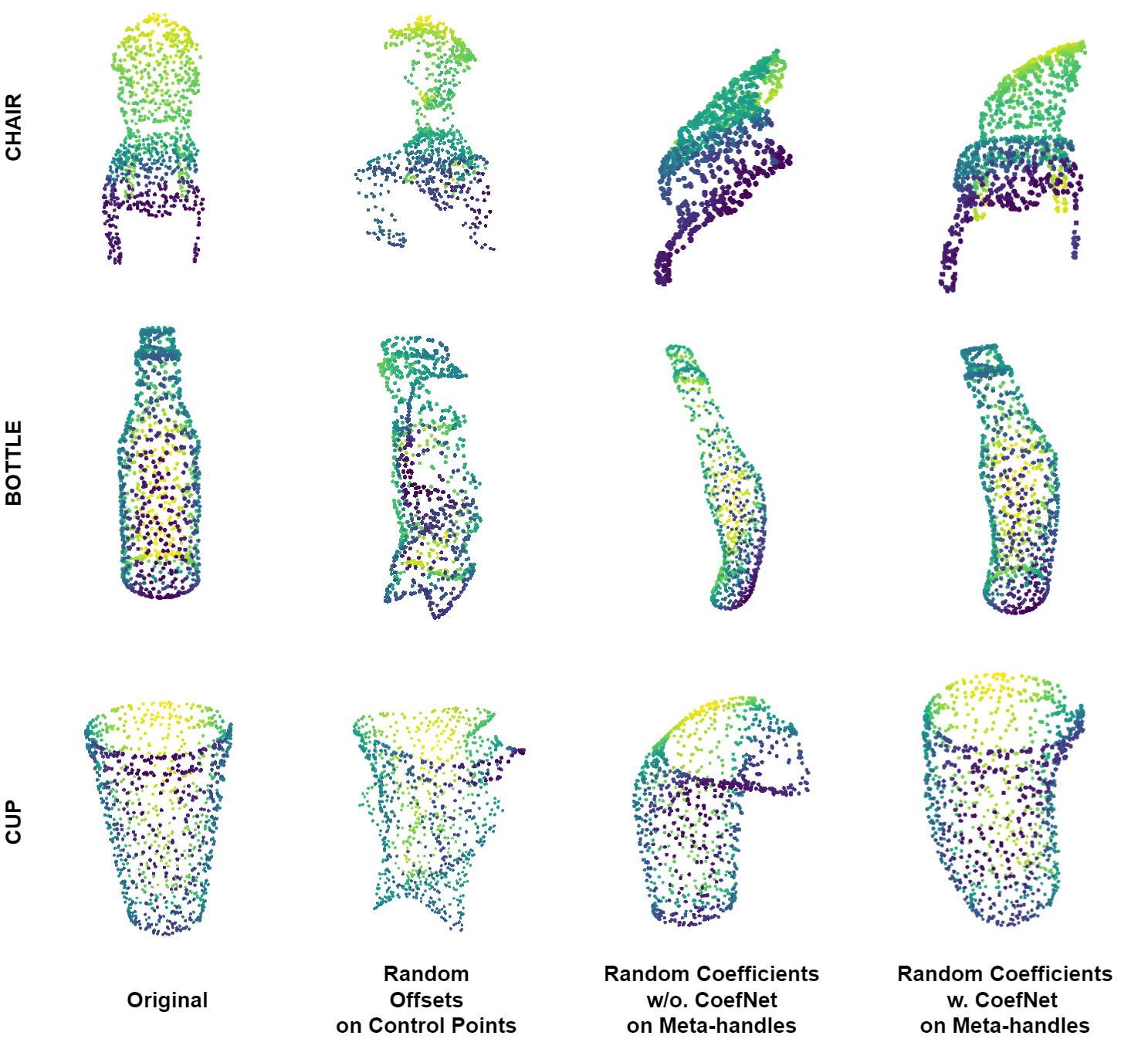}
  \caption{From left to right. We show the original point cloud, the deformed point cloud with random offset draw from $\mathcal{N}(0,0.1)$ on each control points, the deformed point cloud with random coefficients draw from $\mathcal{N}(0,0.5)$ on the deformation prototypes and the deformed point cloud with the same random coefficients but corrected by CoefNet on the deformation prototypes. It is obvious that random offsets on control points will lead to unrealistic and twisted deformations, random coefficients on deformation prototypes would not generate twisted samples but could produce some overwhelming deformation that make the augmented samples unreal and our CoefNet is able to correct those overwhelming deformation and generate realistic deformations.}
  \label{fig:vis2}
\end{figure*}

\subsection{Experiments on CoefNet}
\Cref{tab:rand_a} presents a comparison of PointNet classification performances between random coefficients and coefficients generated by CoefNet. As discussed in \Cref{sec:coefnet}, the coefficients produced by CoefNet are also randomly drawn from normal distributions. However, CoefNet predicts a set of offsets to guide and correct the random coefficients. The results clearly indicate that CoefNet facilitates the generation of more diverse deformations, as it achieves superior performance. Furthermore, the learned coefficients exhibit robustness when the initial coefficients are drawn from distributions with larger variances. Specifically, for $\sigma_a = 1$ and $\sigma_a = 0.7$, the classification performance significantly declines with random coefficients, while the performance remains robust for learned coefficients. This demonstrates CoefNet's capacity to correct undesirable augmentations. \Cref{tab:direct-offset} compares the performance of three backbone networks when CoefNet directly generates the coefficients with input noises and predicts coefficient offsets to guide random coefficients. Predicting offsets to random coefficients outperforms directly predicting coefficients across all three networks, validating our assumption that input noise may not offer sufficient randomness to enhance the diversity of datasets and demonstrating the effectiveness of our designed CoefNet.

\begin{table}[]
\centering
\begin{tabular}{c|ccc}
\hline
                    & PointNet & PointNet++ & DGCNN \\ \hline \hline
Direct coefficient  & 90.8     & 92.0       & 92.9  \\
Coefficient offsets & 91.6     & 93.3       & 93.4  \\ \hline
\end{tabular}%
\caption{Directly predicting coefficients compares to predicting coefficient offsets in CoefNet.}
\label{tab:direct-offset}
\end{table}

\subsection{Ablation Study on AdvTune}
We conduct ablation studies to evaluate the effectiveness of AdvTune. In \Cref{tab:symcd}, we present the classification performance of PointNet trained with and without the two regularization terms. It is evident that both the Chamfer distance and the orientation invariant symmetry loss contribute to the performance of the final result. This substantiates that the two regularization terms can prevent the CoefNet from learning malicious adversarial deformations, thereby mitigating the issue of \textit{distribution mismatch}.

\begin{table}[]
\centering

\begin{tabular}{cc|c}
\hline
$\mathcal{R}_{sym}$ & $\mathcal{R}_{cd}$ & accuracy \\ \hline \hline
     &     & 91       \\
\checkmark    &     & 91.3     \\
     & \checkmark   & 91.3     \\
\checkmark    & \checkmark   & 91.6     \\ \hline
\end{tabular}%

\caption{Ablation study on the two regularization terms Symmetry Loss and Chamfer Distance on PointNet.}
\label{tab:symcd}
\end{table}

\begin{figure*}[]
  \centering

    \includegraphics[width=0.99\textwidth]{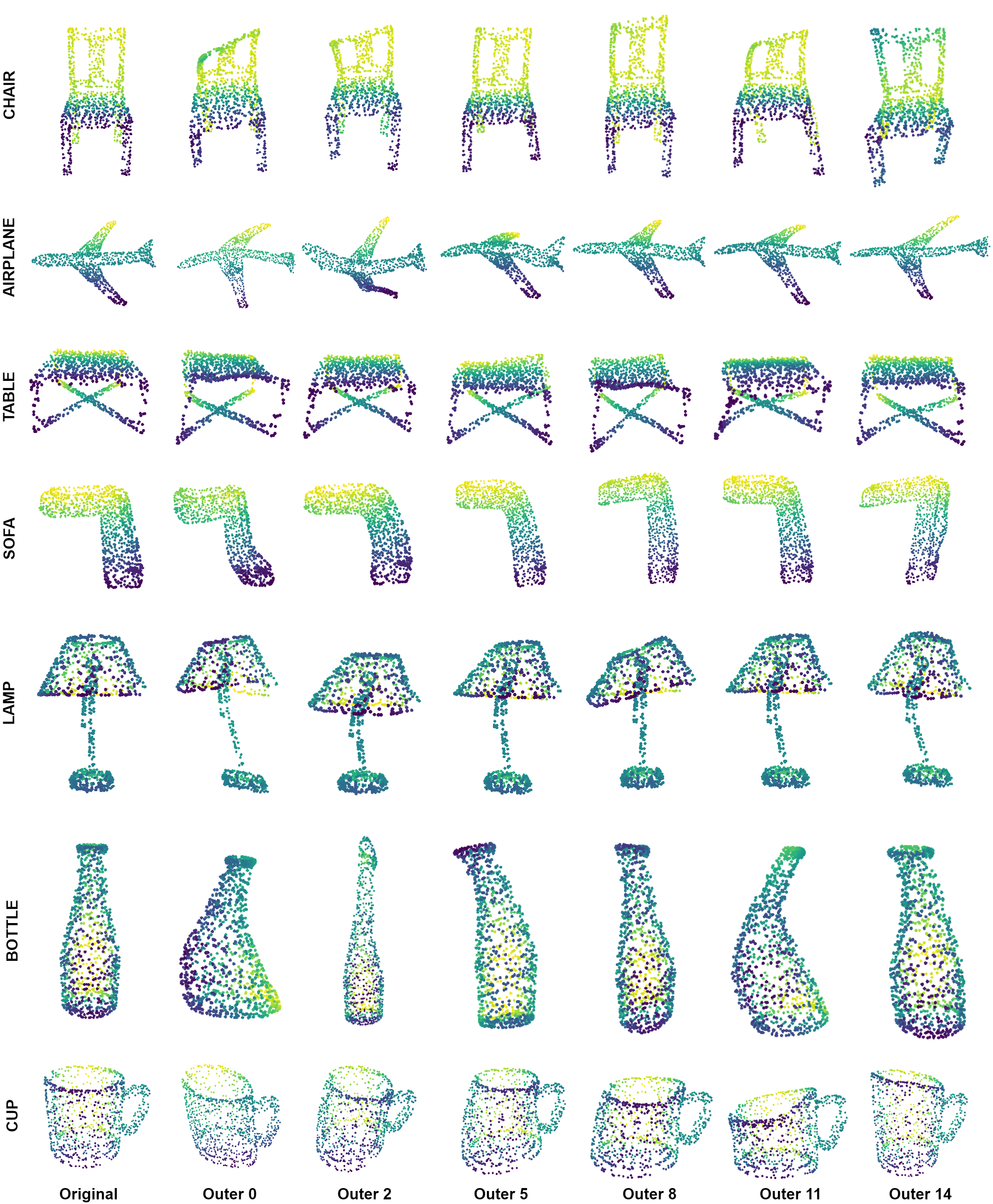}

  \caption{Visualizations of the augmented samples generated by Biharmonic Augmentation in different outer iterations.}
  \label{fig:vis1}
\end{figure*}

\Cref{tab:adv} presents the ablation study of the two fine-tuning modules in AdvTune. DF denotes distribution fine-tuning, corresponding to \Cref{eq:mixture2}, in which the network is trained with clean samples only after training with augmented samples. FF represents final fine-tuning, corresponding to the last loop in \Cref{algorithm2}, where the classification model is fine-tuned after the adversarial training process. The experiments reveal that for PointNet and PointNet++, the classification performance declines considerably without fine-tuning, indicating a severe \textit{distribution mismatch} issue. The performance of PointNet++ is even lower than the baseline result of conventional data augmentation. The second row supports our assumption that solely employing final fine-tuning is inadequate for neutralizing the distribution displacement accumulated during adversarial training, but it can alleviate the effect. The third row demonstrates that distribution fine-tuning, with our designed iterative training scheme, effectively reduces distribution displacement during adversarial training. It yields improvements of +2.6\%, +4.7\%, and 1.1\% for PointNet, PointNet++, and DGCNN compared to the first row results without fine-tuning. In each outer loop, distribution fine-tuning is performed immediately after training with augmented samples, where distribution displacement occurs. Subsequently, CoefNet training follows distribution fine-tuning, which implies that displacement does not accumulate during the training. The fourth row indicates that, with the iterative training scheme, final fine-tuning serves as a complementary module that brings minor improvements to the final results. Thus, the experiments demonstrate that our AdvTune plays a significant role in training both the classification network and the CoefNet.

\begin{table}[htb]
\centering
\begin{tabular}{c|ccc}
\hline
Training   & PointNet & PointNet++ & DGCNN \\ \hline \hline
w/o DF+FF & 88.9     & 87.0       & 92.3  \\
FF w/o DF & 90.5     & 91.7       & 92.5  \\
DF w/o FF & 91.5     & 93.0       & 93.4  \\
DF + FF   & 91.6     & 93.3       & 93.4  \\ \hline
\end{tabular}
\caption{Ablation study on AdvTune on three classification networks. DF denotes distribution fine-tuning and FF denotes final fine-tuning.}
\label{tab:adv}
\end{table}

\begin{table}[htb]
\centering
\begin{tabular}{c|ccc}
\hline
Corruption   & w/o DA & CDA  & BA \\ \hline
jitter $\sigma=0.01$  & 87.1   & 88.1 & 91.0      \\
jitter $\sigma=0.03$   & 84.3   & 83.5 & 88.6      \\
jitter $\sigma=0.05$   & 66.2   & 62.8 & 72.6      \\ \hline
Dropout 0.25 & 87.8   & 88.3 & 91.2      \\
Dropout 0.5  & 86.5   & 87.5 & 90.9      \\
Dropout 0.75 & 80.0   & 78.2 & 84.6      \\ \hline
Scale 0.9    & 85.8   & 88.4 & 91.6      \\
Scale 1.1    & 86.3   & 88.9 & 91.3      \\ \hline
\end{tabular}%

\caption{Robustness test on PointNet with different corruptions in the testing data.}
\label{tab:corruption}
\end{table}

\subsection{Robustness Test}
As shown in \Cref{tab:corruption}, we perform a robustness test to our model with corrupted testing data. The experiments are performed on PointNet. We corrupt the testing data with different levels of random jittering, random dropout and scaling. And we compare our method with models trained with no data augmentation and conventional data augmentations. For random jittering, the performance difference is larger when the jittering gets stronger. Conventional data augmentation even performs worse than the model trained with no data augmentation. Our method also shows robustness to random dropout and scaling compared to conventional data augmentation.

\subsection{Qualitative Results}\label{sec:quali}
\Cref{fig:vis2} compares four types of point clouds: the original point cloud, the deformed point cloud with random offsets drawn from $\mathcal{N}(0,0.1)$ on each control point, the deformed point cloud with random coefficients drawn from $\mathcal{N}(0,0.5)$ on the deformation prototypes, and the deformed point cloud with the same random coefficients but corrected by CoefNet on the deformation prototypes. As shown in the second column of \Cref{fig:vis2}, randomly moving individual control points can easily create distorted deformations that are difficult for humans to recognize their correct categories. This issue is resolved by using deformation prototypes. However, although deformation prototypes with random coefficients do not create distorted deformations, they may generate excessive deformations on specific prototypes. As illustrated in the third column of \Cref{fig:vis2}, overwhelming deformation can also produce unrealistic deformations. In contrast, when we feed the same random coefficients to our proposed CoefNet, as shown in the fourth column of \Cref{fig:vis2}, we can correct the excessive deformation and generate realistic deformations. The experiment results in \Cref{tab:rand_a} show that performance declines with random coefficients when they are drawn from a distribution with a larger standard deviation, which experience the same issue illustrated in the third column of \Cref{fig:vis2}.

\Cref{fig:vis1} presents additional visualizations of augmented samples at different training stages. We can observe that the deformed shapes vary during the training, indicating that the deformations do not converge to specific shapes, and thus providing sufficient diversity throughout the training process.

This demonstrates that our proposed CoefNet is capable of correcting the random coefficients to produce realistic deformations while preserving the diversity offered by the random coefficients.

\section{Future works}
The proposed method exhibits potential for future extensions, particularly towards scene-level applications. A plausible avenue for advancement involves augmenting 3D scenes by manipulating individual object shapes within the scene. Adversarial training can be leveraged to identify optimal augmentations that enhance performance in scene-understanding tasks. Additionally, as a part of prospective research, exploration into alternative non-rigid deformation techniques presents an opportunity to expedite the training process for deformation prototype networks. These prospective endeavors hold promise for further enhancing the capabilities and efficiency of the method.

\section{Conclusion}

This paper introduces Biharmonic Augmentation, a novel and impactful data augmentation approach designed for point cloud classification tasks. Our method achieves smooth and realistic deformations on the original data, effectively expanding the dataset. These deformations are efficiently generated through straightforward matrix multiplication techniques. Additionally, we present AdvTune, an online auto-augmentation framework empowered by adversarial training, which empowers the augmentor network CoefNet to generate more diverse and meaningful augmentations tailored to the target network. The utilization of deformation-based data augmentation offers significant potential, addressing the data scarcity challenge in 3D point cloud data by producing authentic and expanded dataset augmentations.

 {\small
\bibliographystyle{IEEEtran}
\bibliography{IEEEfull}
}
\newpage

\vfill

\end{document}